\documentclass[10pt, a4paper]{article}

\usepackage{lrec-coling2024} 
\usepackage{comment}
\usepackage{listings}
\usepackage{subcaption}
\usepackage{natbib}
\usepackage{multibib}
\makeatletter
\def\@mb@citenamelist{cite,citep,citet,citealp,citealt,citepalias,citetalias}
\makeatother
\newcites{languageresource}{~}

\usepackage{graphicx}
\usepackage{tabularx}
\usepackage{soul}
\usepackage{amsmath}
\usepackage{mdframed}
\usepackage{xcolor}
\usepackage{listings}

\usepackage{titlesec}
\titleformat{\section}{\normalfont\large\bfseries\center}{\thesection.}{1em}{}
\titleformat{\subsection}{\normalfont\SmallTitleFont\bfseries\raggedright}{\thesubsection.}{1em}{}
\titleformat{\subsubsection}{\normalfont\normalsize\bfseries\raggedright}{\thesubsubsection.}{1em}{}
\renewcommand\thesection{\arabic{section}}
\renewcommand\thesubsection{\thesection.\arabic{subsection}}
\renewcommand\thesubsubsection{\thesubsection.\arabic{subsubsection}}

\usepackage{xcolor}
\usepackage{hyperref}
 \definecolor{darkblue}{rgb}{0, 0, 0.5}
  \hypersetup{colorlinks=true, citecolor=darkblue, linkcolor=darkblue, urlcolor=darkblue}

\usepackage{xstring}

\usepackage{color}
\definecolor{mypurple}{rgb}{0.5,0,0.5}

\lstdefinestyle{pythonstyle}{
    backgroundcolor=\color{pink},   
    basicstyle=\ttfamily\fontsize{7pt}{10pt},      
    commentstyle=\color{mypurple},    
    keywordstyle=\color{mypurple},       
    numberstyle=\tiny\color{mygray}, 
    stringstyle=\color{mypurple},    
    numbers=left,                    
    stepnumber=1,                    
    numbersep=5pt,                   
    breakatwhitespace=false,         
    breaklines=true,                 
    frame=single,                    
    captionpos=b,                    
    numbers=none,
    tabsize=4                        
}
\definecolor{mycyan}{rgb}{0,0.6,0.6}
\lstdefinestyle{plaintextstyle}{
    backgroundcolor=\color{lightgray},   
    basicstyle=\ttfamily\fontsize{7pt}{10pt}\selectfont\color{violet}, 
    numbers=none,                    
    frame=none,                      
    breakatwhitespace=false,         
    breaklines=true,                 
    tabsize=4                        
}

\title{CLEVR-POC: Reasoning-Intensive Visual Question Answering in Partially Observable Environments}

\name{Savitha Sam Abraham$^{\ast}$, Marjan Alirezaie$^{\dagger}$, Luc De Raedt$^{\S}$}

 \address{$^{\ast}$The University of Adelaide, Australia \\ 
          {savitha.samabraham@adelaide.edu.au} \\ \\
          $^{\dagger}$Flybits Labs. TMU Creative AI Hub, Toronto, Canada \\ 
          marjan.alirezaie@flybits.com \\ \\
          $^{\S}$Örebro University, Centre for Applied Autonomous Sensor Systems(AASS), Örebro, Sweden\\
          Department of Computer Science, KULeuven, Belgium\\
          luc.de-raedt@oru.se\\ 
        }

\abstract{
The integration of learning and reasoning is high on the research agenda
in AI. Nevertheless, there is only a little attention
to use existing background knowledge for reasoning about partially observed scenes to answer questions about the scene. Yet, we as humans use such knowledge frequently to infer plausible answers to visual questions (by eliminating all inconsistent ones). Such knowledge often comes in the form of 
constraints about objects and it tends to be highly domain or environment-specific. We contribute a novel benchmark called CLEVR-POC\footnote{POC stands for Partial Observability with Constraints.} for reasoning-intensive visual question answering (VQA) in partially observable environments under constraints. In CLEVR-POC, knowledge in the form of logical constraints needs to be leveraged to generate \textit{plausible answers} to questions about a hidden object in a given partial scene. For instance, if one has the knowledge that all cups are colored either red, green or blue and that there is only one green cup, it becomes possible to deduce the color of an occluded cup as either red or blue, provided that all other cups, including the green one, are observed. Through experiments, we observe that the low performance of pre-trained vision language models like CLIP ($\approx$ 22\%) and a large language model (LLM) like GPT-4 ($\approx$ 46\%) on CLEVR-POC ascertains the necessity for frameworks that can handle reasoning-intensive tasks where environment-specific background knowledge is available and crucial. Furthermore, our demonstration illustrates that a neuro-symbolic model, which integrates an LLM like GPT-4 with a visual perception network and a formal logical reasoner, exhibits exceptional performance on CLEVR-POC. 
\newline  
}

\begin{document}

\maketitleabstract

\section{Introduction}
Visual Question Answering (VQA) has been widely investigated by researchers from various subfields in AI like computer vision and natural language understanding. As a result, we now have access to a vast corpus of VQA datasets coupled with numerous models addressing the task of VQA \cite{zou2020survey, wu2017visual}.

\begin{figure*}
\centering
\begin{subfigure}{\textwidth}
    \includegraphics[width=\textwidth,height=4cm]{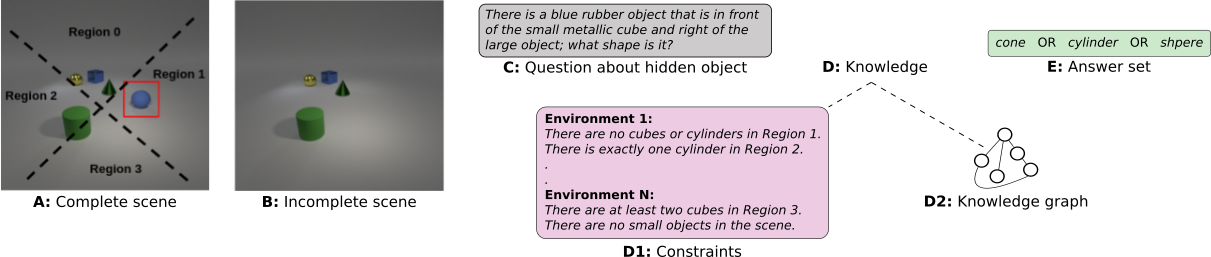}
    \caption{ The different components in VQA tasks.}
    \label{fig:first}
\end{subfigure}
\hfill
\begin{subfigure}{\textwidth}
    \includegraphics[scale = 0.3]{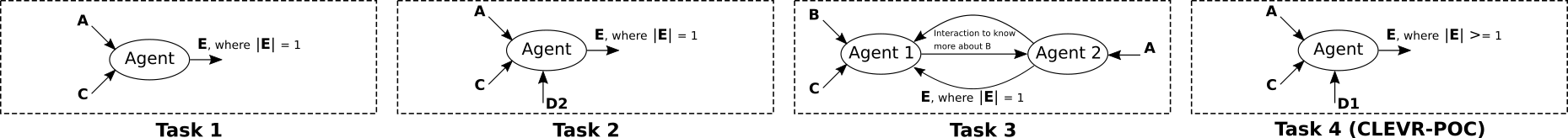}
    \caption{The different VQA tasks are based on expected inputs and outputs and the number of agents involved. }
    \label{fig:third}
\end{subfigure}
\caption{VQA task components and types of VQA tasks }
\label{fig:figures}
\end{figure*}

 Most existing VQA datasets \cite{johnson2017clevr, antol2015vqa} have a collection of images paired with questions such that all information required to answer the question is provided in the image, and hence the scene is considered complete.  
 But in real life, we often engage in tasks where scenes may not be completely visible. We instead may have world knowledge about various locations visited by us, acquired over time, that allows us to generate plausible answers to queries about objects we do not see in a scene.  
For example, in autonomous vehicle scenarios, reasoning is crucial for dealing with partial observability. Comprehensive knowledge of traffic enables the system to interpret limited visual information and make informed decisions, ensuring safe navigation despite occlusions or limited field of view. Furthermore, in factory settings, reasoning combined with background knowledge about the environment can assist teams of robots in dealing with partial observability during navigation and other coordination and cooperation tasks. 

In this paper, we introduce a synthetic dataset, CLEVR-POC\footnote{The source code associated with this research project is openly accessible at \url{https://github.com/savithasam88/CLEVR-POC/tree/master}}, for a reasoning-intensive VQA task set in partially observable scenarios involving external knowledge in the form of constraints. The dataset consists of pairs of an image, representing a partial scene (\textbf{B} in Figure \ref{fig:first}) in some environment (\textbf{D1} in Figure \ref{fig:first} where the environment is defined by a set of constraints), and a question in natural language about some hidden/missing object (\textbf{C} in Figure \ref{fig:first}) in the scene. Although in the literature, there exist datasets for QA tasks in partially observable environments (e.g., CLEVR-dialog \cite{kottur2019clevr}, Visual Dialog \cite{das2017visualD}, Guess What? \cite{de2017guesswhat}), these do not come with additional background knowledge. The challenge presented in CLEVR-POC necessitates the integration of knowledge and multi-step reasoning involving eliminative induction, into perception systems driven by learning. Given that the knowledge associated with a scene typically varies depending on the specific environment involved, it is not a constant across the dataset. It becomes challenging for a learning system to simply memorize this knowledge during training iterations. Moreover, because this knowledge is environment-specific, employing Large Language Models (LLMs) such as GPT as the source of knowledge, as demonstrated in some of the recent works like \cite{zhou2023navgpt} and \cite{shah2023navigation}, does not yield favorable results. We substantiate these assertions through empirical experiments. 

The contributions of this paper are as follows:

\begin{itemize}
    \item We introduce a dataset, CLEVR-POC, that introduces the task of reasoning-intensive VQA - given a \textit{partial scene}, the \textit{constraints} (knowledge) to which the scene conforms and a \textit{question} about a hidden object in the scene, find the set of all plausible answers.
    \item We evaluate the performance of state-of-the-art pre-trained vision language and large language models on CLEVR-POC. 
    \item We demonstrate that the synergistic use of LLMs alongside a visual perception network and a formal reasoning system with access to external knowledge can efficiently and effectively address the challenges presented by CLEVR-POC.     
\end{itemize}

The organization of the paper is as follows. Section \ref{sec:rel} provides an overview of existing work in VQA, focusing on various VQA datasets and briefly discussing LLM for reasoning. Section \ref{sec:approach} delves into the detailed process of generating CLEVR-POC, while Section \ref{sec:experiments} outlines the research questions explored in this study. Additionally, this section presents the experiments conducted on CLEVR-POC and the corresponding results.         

\section{Related Work} \label{sec:rel}
In this section, we provide an overview of research in two domains -  datasets in VQA and LLMs and reasoning.  

\subsection{Datasets in VQA}

A VQA task may involve various combinations of the different components shown in Figure \ref{fig:first} - a complete scene (\textbf{A}), a partial scene (\textbf{B}), a question (\textbf{C}) about the scene, external knowledge in the form of rules/constraints (\textbf{D1}), or facts in knowledge graphs (\textbf{D2}), and the set of plausible answers to the question (\textbf{E}). Each combination results in a different VQA task (see Figure \ref{fig:third}). 

\subsection{Types of VQA Tasks}
\label{sec:tasks}

    \subsubsection {\textbf{Task 1}} Given a \textit{complete scene}, and a \textit{question} about an object in the scene, find the answer to the question. Since the scene is complete, the agent can come up with the exact answer implying that the solution set \textbf{E} has just one element ($|$\textbf{E}$|$ = 1). DAQUAR \cite{malinowski2014multi}, VQA \cite{antol2015vqa}, CLEVR \cite{johnson2017clevr} are datasets in this category.   
    \subsubsection {\textbf{Task 2}} Given a \textit{complete scene}, a \textit{question} about one of the objects in the scene and external knowledge about objects (in the form of triples - \textbf{D2}), find the answer to the question leveraging this external knowledge. FVQA (fact-based VQA) \cite{wang2017fvqa}, and KVQA (knowledge aware VQA) \cite{Shah_Mishra_Yadati_Talukdar_2019} are datasets in this category.
    \subsubsection{ \textbf{Task 3}} While the above two tasks involve a single agent being posed with a scene and a question, this category of VQA tasks involves more than one agent. One of the agents has access to the complete scene while the other agent is provided with a partial scene and a question. Answering the question requires the agents to interact with each other. CLEVR-dialog \cite{kottur2019clevr}, Visual Dialog \cite{das2017visualD}, Guess What? \cite{de2017guesswhat} are datasets handling Task 3.
    \subsubsection{ \textbf{Task 4 (CLEVR-POC)}} Given a \textit{partial scene}, \textit{knowledge in the form of rules (constraints)} about the environment to which the scene conforms and a \textit{question} about the hidden object in the scene, find the set of all plausible answers to the question. Since the question is about a hidden object (for example, about the shape of the object), it may not be always possible to provide an exact solution. Answering the question is more about eliminating all cases that are inconsistent with the background knowledge (for example: given the knowledge -  \textit{there are no spheres in this environment}) and returning all consistent answers as the solution (\textit{the shape is a cone or a cylinder or a cube}, which is why $|\textbf{E}|$ $\geq$ 1). In contrast to Task 2, where the knowledge graph encompasses general world facts (e.g.,``\textit{cows are herbivores}"), the knowledge in this context is considerably more specific to an environment. While an LLM can be presumed to possess awareness of the former category of knowledge, the same cannot be said for the latter.

\subsection{LLMs and Reasoning}
In this paper, our emphasis lies on the process of reasoning which depends on a formal system grounded in logical rules and principles. Such a system ensures that all transformations or manipulations of symbols within it, leading to new inferences, adhere to the logical rules and principles governing the system \cite{maccoll1897symbolic}.  While LLMs can also be seen as performing symbolic manipulations, these manipulations unlike traditional symbolic reasoning are based on statistical associations or patterns learned from data \cite{huang-chang-2023-towards}, because of which it may or may not be logically sound. Despite the progress in the development of large language models (LLMs), many still struggle with a deep understanding of symbols like humans do \cite{abraham2022compositional, yao2022react}. To address this gap, there are ongoing efforts to create benchmarks \cite{huang-chang-2023-towards}, like the proposed CLEVR-POC, to evaluate the reasoning capabilities of LLMs. 

In CLEVR-POC, we introduce a VQA task that involves constraint-based reasoning, a form of logical reasoning, where the generated response must satisfy a set of constraints given. These benchmarks are used to assess the capacity of language models in handling symbolic reasoning, contributing to the advancements in the development of more logically sound systems.

\begin{table*}[]
\centering
\small
\begin{tabular}{|p{0.8\textwidth}|p{0.1\textwidth}}
\cline{1-1}
{\bf Template-1 (Value Restriction)}  \\
\rmfamily{:- object(X),at(X, R'), not hasProperty(X, P1', V1').} \\
\textit{Translation} All objects at region \textit{R'} have value \textit{V1'} for the property \textit{P1'}. \\
\textit{An instantiation} \rmfamily{:- object(X),at(X, 0), not hasProperty(X, color, red).} &\\
\cline{1-1}
\textbf{Template-2 (Negation Constraint)} \\
\rmfamily{:- object(X), at(X, R'), hasProperty(X, P1', V1').} \\
\textit{Translation} All objects at region \textit{R'} cannot have value \textit{V1'} for the property \textit{P1'}. \\
\textit{An instantiation} \rmfamily{:- object(X), at(X, 0), hasProperty(X, material, metal).} &\\
\cline{1-1}
\textbf{Template-3 (Exactly N Constraint)} \\
\rmfamily{:- $\#$count$\{$X: hasProperty(X, P1', V1'), object(X), at(X, R')$\}$!=N'} \\
\textit{Translation} There are exactly \textit{N'} objects at region \textit{R'} with value \textit{V1'} for the property \textit{P1'}. \\
\textit{An instantiation} \rmfamily{:- $\#$count$\{$X: hasProperty(X, size, small), object(X), at(X, R')$\}$!=2} &\\
\cline{1-1}
\textbf{Template-4 (Atleast N Constraint)} \\
\rmfamily{:- $\#$count$\{$X1, X2: sameProperty(X1, X2, P1'), object(X1), object(X2), at(X1, R1'), at(X2, R2')$\}$ $<$ N'. }\\
\textit{Translation} There are at least $N'$ pairs of objects at regions  \textit{R1'} and \textit{R2'} that has the same value \textit{V1'} for the property \textit{P1'}. \\
\textit{An instantiation} \rmfamily{:- $\#$count$\{$X1, X2: sameProperty(X1, X2, shape), object(X1), object(X2), at(X1, 1), at(X2, 2)$\}$$<$1.} &\\
\cline{1-1}
\textbf{Template-5 (OR Constraint)} \\
\rmfamily{:- object(X), at(X, R'), not hasProperty(X, P1', V1'),  not hasProperty(X, P1', V2').} \\
\textit{Translation}  All objects in region \textit{R'} have value \textit{V1'} for property \textit{P1'} or \textit{V2'} for property \textit{P2'}. \\
\textit{An instantiation} \rmfamily{:- object(X), at(X, 1), not hasProperty(X, color, yellow),  not hasProperty(X, color, blue).} &\\
\cline{1-1}
\end{tabular}
\caption{A few constraint templates}
\label{tab:constraints}
\end{table*}

\section{The CLEVR-POC Dataset } \label{sec:approach}
Now we describe in detail the generation of the CLEVR-POC dataset. The dataset, as the name suggests, is based on the CLEVR \cite{johnson2017clevr} dataset, which generated scenes with geometrical shapes. Each object is associated with four attributes - color, shape, material, and size. The objects in CLEVR-POC can have one of the four shapes - cone, sphere, cylinder, and cube, three sizes - large, medium, and small, two materials - rubber and metal, and eight colors - red, blue, green, yellow, gray, brown and purple. Besides these four attributes, since a scene is divided into four regions (see Figure \ref{fig:first}), CLEVR-POC also associates an object with the region it is in - $0, 1, 2$ or $3$. Each object belongs to exactly one region. Division of a scene into regions enables the specification of constraints at multiple levels.
\begin{itemize}
    \item \textbf{Region-based} constraints - for example, all objects in Region $0$ are of shapes cube or cylinder. These constraints must be satisfied by objects in the corresponding region.
    \item \textbf{Across-region} constraints - for example, the total number of objects sharing the same color in regions 1 and 2 is not more than 2. These are constraints specified across two regions.  
    \item \textbf{Generic} constraints - for example, there are at least two cubes in the scene. These constraints apply to the whole scene. 
\end{itemize}

One of the major points of distinction in the scene generation process of CLEVR-POC from the original CLEVR is that the scenes in CLEVR-POC are generated such that they conform to a chosen set of constraints. The steps in creating an instance $i$ in the dataset are:
\begin{itemize}
    \item Generating an environment - $Environment_i$, defined by a set of constraints.
    \item Generating a complete scene graph, $Complete_i$, that conforms to $Environment_i$.
    \item Generating the partial scene graph, $Partial_i$ by removing one of the objects, $Obj_i$, from $Complete_i$.
    \item Generating a question, $Q_i$, about the partial scene with object of interest $Obj_i$.  
\end{itemize}

\subsection{Environment Representation}
An environment in CLEVR-POC is defined by a set of constraints. We provide a set of 11 constraint templates with CLEVR-POC that are expressed in answer set programming (ASP)\footnote{ASP is a declarative programming paradigm applied to solve complex search problems~\cite{10.5555/1620270.1620340}}. Each environment is created by at most 15 different instantiations of these templates, provided there are at least two constraints associated with each region. A few example constraint templates with their translation in English and an instantiation are shown in Table \ref{tab:constraints}. Around 30 different environments are generated (see Appendix A for an example) and the scenes in the dataset belong to one of these 30 environments - the dataset generation process ensures that the scenes are uniformly distributed across the 30 environments.

\begin{figure*}
\centering
\begin{subfigure}{\textwidth}
    \includegraphics[scale = 0.18]{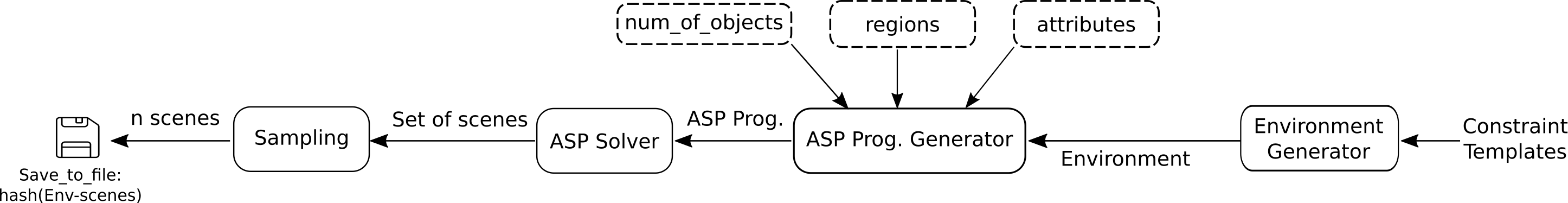}
    \caption{Pipeline for generating \textit{environment} and \textit{complete scenes} in that environment.}
    \label{fig:2first}
\end{subfigure}
\hfill
\begin{subfigure}{\textwidth}
    \includegraphics[scale=0.165]{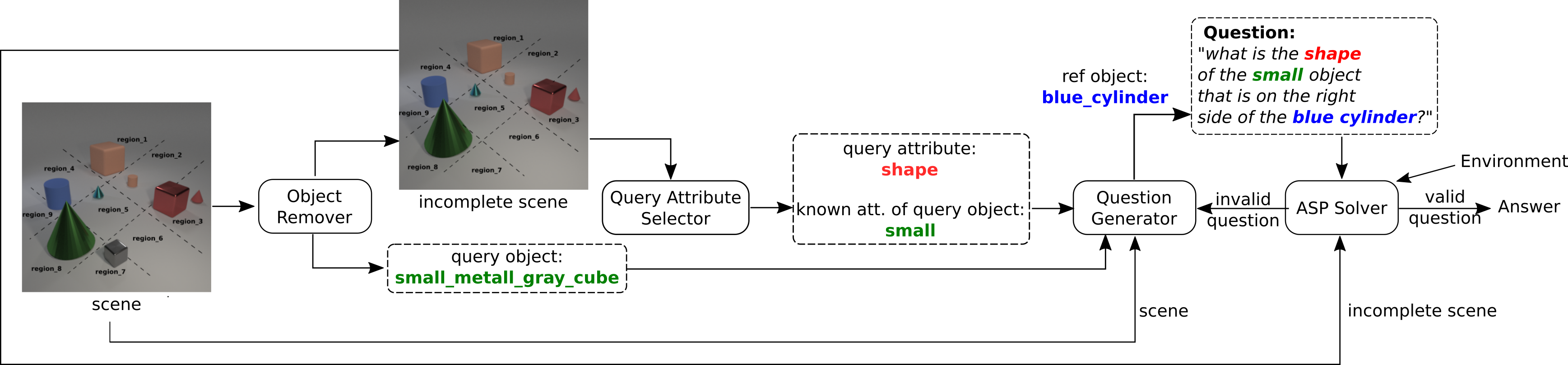}
    \caption{Pipeline for generating partial scenes, and questions and then labeling them with answers.}
    \label{fig:2third}
\end{subfigure}
        
\caption{Two steps in dataset generation process: Figure \ref{fig:2first} shows the first step - environment generation from constraint templates and generating complete scenes satisfying these constraints. Figure \ref{fig:2third} shows Step 2 - partial scene and question generation from a complete scene.}
\label{fig:figures-1}
\end{figure*}

\subsection{Scene Representation}
CLEVR represented a scene in the form of a scene graph whose nodes represented objects annotated with its attributes and edges denoted the spatial relations (left, right, front, behind) between objects. In CLEVR-POC, besides the scene graph representation, we also represent a scene in ASP. Below we show part of the ASP representation of the partial scene in Figure \ref{fig:first}.  
\begin{small}
\begin{verbatim}
%Objects in the scene
object(0). object(1). object(2). object(3).

%Attributes of objects  
at(0, 2).
hasProperty(0, color, green).
hasProperty(0, size, large).
hasProperty(0, material, rubber).
hasProperty(0, shape, cylinder).
....
%Spatial relations between objects 
front(1, 0).   right(1, 0). ...
\end{verbatim}
\end{small}
The predicate \texttt{object} is used to define the different objects (denoted using identifiers - 0, 1, ..). \texttt{hasProperty(Object, Attribute, Value)} associates a \texttt{Value} for an \texttt{Attribute} of an \texttt{Object}. \texttt{at(Object, Region)} represents the region where the object is located. The spatial relations between objects are represented with predicates \texttt{left, right, front, behind} - for example, \texttt{left(Object1, Object2)} represents that \texttt{Object2} is located left of \texttt{Object1}. 

\subsection{Image Generation}
While the images in CLEVR are generated from a randomly sampled scene graph, CLEVR-POC generates its images from scene graphs known to adhere to constraints defining an environment. Scene graph creation is thus a reasoning problem - given an environment (constraints in ASP) and a desired number of objects (\textit{n}) in the scene, the goal is to assign each object to one of the four regions and propose values to color, size, shape, and material that are consistent with the constraints in the environment. An ASP reasoning engine solves this problem - each answer (a consistent property-value assignment for the \textit{n} objects) in the answer set returned is a scene graph or a possible configuration of the objects in the scene. Since there are many possible configurations - we randomly sample a million of these scene graphs for the subsequent image generation phase. A scene graph is then rendered using Blender\footnote{\url{https://www.blender.org/}}. The image representing the partial scene is generated from a partial scene graph constructed from the actual scene graph by randomly removing one of the objects from it. Figure \ref{fig:2first} shows the scene graph construction process.

\subsection{Question Representation}\label{quest_repr}
The questions in CLEVR-POC query about one of the four attributes - color, size, shape, and material of the missing/hidden object in the partial scene. Besides representing the questions using an equivalent functional program as in CLEVR, CLEVR-POC also represents it in ASP.  An example question and its ASP form are shown below:

\begin{quote}
\small
\textbf{Question}: \\
\textit{What is the color of the other cylinder that is the same material as the medium red thing? } \\\\
\vspace{-1.0ex}
\rmfamily{\texttt{query(Q):- hasProperty(X,color,Q),}} \\
\hspace*{1.9 cm} \vspace{0.1cm} \rmfamily{\texttt{hasProperty(X,shape,cylinder),}} \\ \hspace*{1.9cm}  \rmfamily{\texttt{hasProperty(Y,size,medium),}} \\ \hspace*{1.9cm} \rmfamily{\texttt{hasProperty(Y,color,red),}} \\
     \hspace*{1.9cm} \rmfamily{\texttt{same\_material(Y,X),}} \\
     \hspace*{1.9cm} \rmfamily{\texttt{X!=Y.}} 
\end{quote}

If the query is about attribute $A$, $A \in \{color, size, material, shape\}$, the questions are generated such that the cardinality of the set of possible solutions ($S$) is  $1 \leq |S| < |A|$,  where $|A|$ is the set of all values for the attribute $A$ (for example $|size|$ = 3 = $|\{large, medium, small\}|$). If the question generated has $|A|$ solutions (for instance, a solution like, `\textit{size is large or small or medium}' is true for any question), it is considered invalid. 
The questions are balanced across the question types (that depend on the query attribute - see Appendix B for the distribution). It should be noted that the solution space of CLEVR-POC questions is 16 times that of CLEVR as the solutions expected are not always a single value, but a set of values.

\subsection{Question Generation}
The question in CLEVR-POC is generated from the question templates available in CLEVR. We avoid the yes/no (existence, comparison) and counting questions and focus on just the attribute querying templates. An example template is as follows:
\begin{quote}
\small
What shape is the $<Z2>(size) <C2>(color) <M2>(material)$ [that is] $<R>(relation)$ the $<Z>(size) <C>(color) <M>(material) <S>(shape)?$\\
\end{quote}
\vspace{-3ex}
 Question template instantiation is done based on the complete scene graph of the associated image. The object of interest is always the object that is removed from the complete scene to generate the partial scene graph. The query attribute is picked such that it satisfies the question type balancing requirements. The known attributes of the query object (filling the slots $<Z2>$ or $<C2>$ or $<M2>$ in the above template) are randomly selected. While the filler for the slot $<R>$ (one of the left, right, front, behind) is randomly picked, the reference object in the query is picked based on the spatial relations available in the complete scene - picking one of the objects that are in $<R>$ relation of the query object. 

The ASP representations of the question, the incomplete scene, and the constraints in the environment are given to an ASP solver to identify the set of possible values for the query attribute. Figure \ref{fig:2third} shows the pipeline of question generation. Refer to Appendix A and B for a detailed example and statistics of CLEVR-POC.     

\begin{figure*}
\centering
  \includegraphics[scale = 0.18]{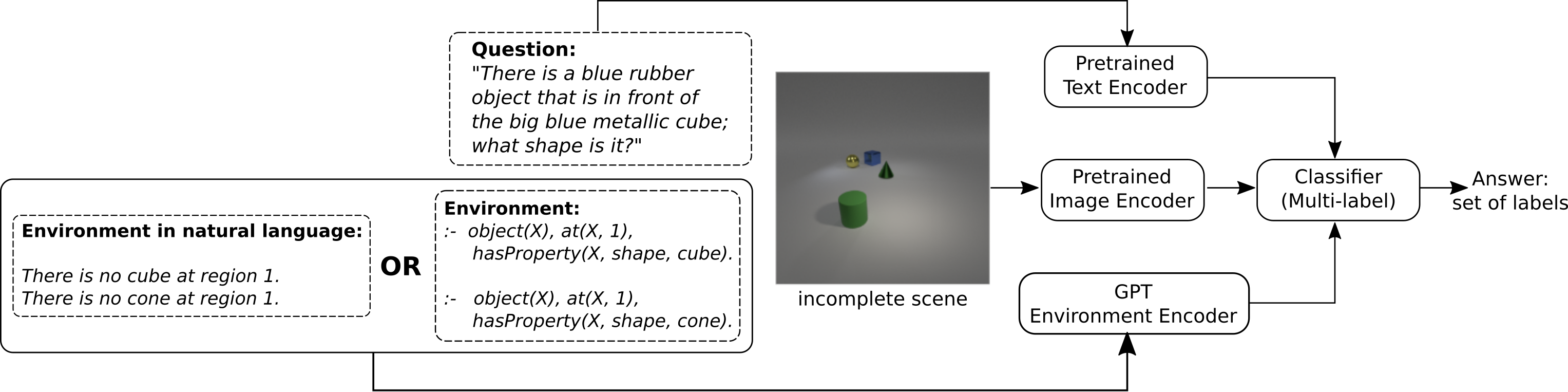}
  \caption{CLIP for CLEVR-POC}
\label{fig:CLIP}
\end{figure*}
\section{Experiments} 
\label{sec:experiments}

The experiments are designed to answer the following research questions (RQ):
\begin{itemize}
    \item \textbf{RQ1}: How do neural-based vision language models perform on reasoning-intensive VQA tasks (with an emphasis on symbolic knowledge representation and reasoning)?
    \item \textbf{RQ2}: How well do neuro-symbolic vision language architectures handle reasoning-intensive VQA tasks (in the context of mapping raw inputs to symbolic space)?
    \item \textbf{RQ3}: How can we leverage LLMs in reasoning-intensive VQA tasks and what are the challenges associated with it?
\end{itemize}

In the sections following, we describe the methods implemented to answer these questions. 
\subsection{Methods}

\begin{figure*}
\centering
  \includegraphics[scale = 0.18]{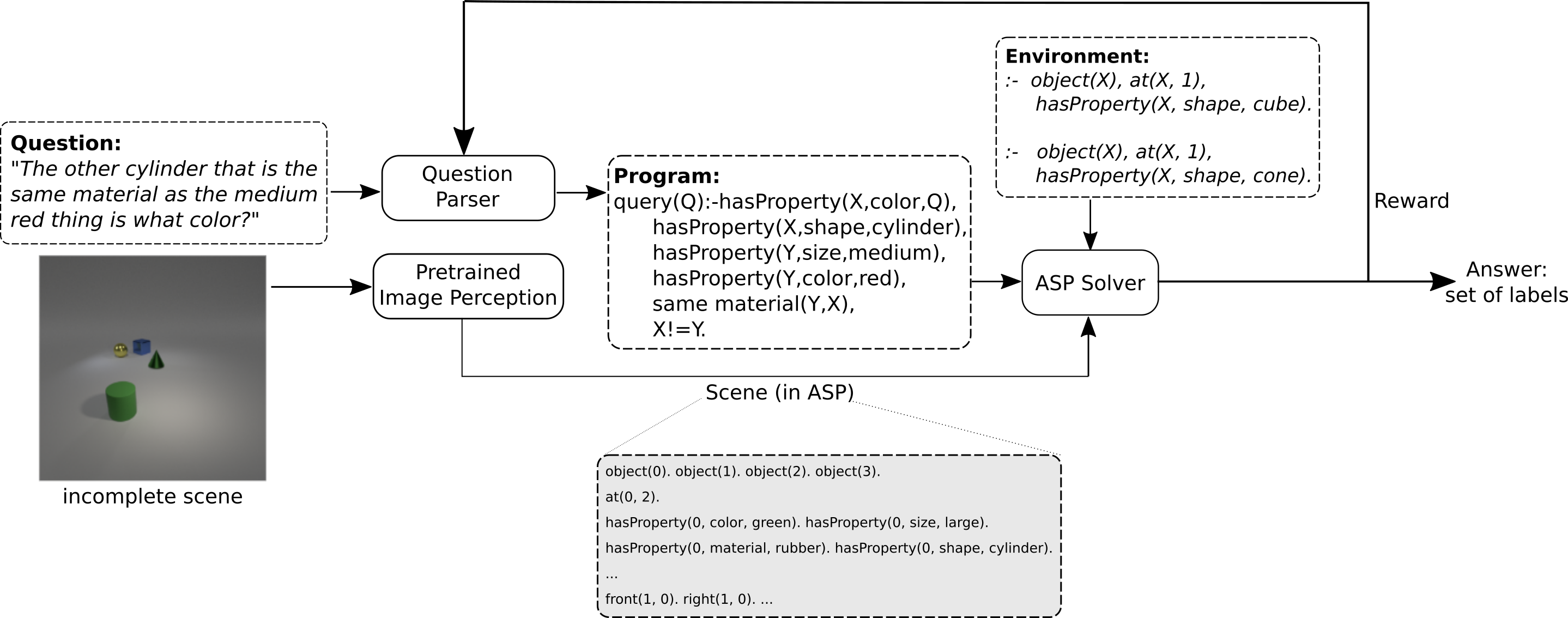}
  \caption{NS-VQA for CLEVR-POC - architecture is updated with an ASP solver}
\label{fig:NSVQA}
\end{figure*}

\subsubsection{CLIP-based model}
 CLIP (Contrastive Language Image Pre-training) \cite{radford2021learning} is a vision-language model that is trained to align pairs of text and images to a unified space. We experimented with the CLIP model to investigate RQ1. Figure \ref{fig:CLIP} shows the architecture of a CLIP-based model to solve CLEVR-POC. 
 The pre-trained vision transformer (ViT-B/32) and the text encoders (masked self-attention) in CLIP are leveraged to obtain encodings for the incomplete scene and the question. The encoding for the environment is obtained from its constraints. A pre-trained GPT-2 \cite{radford2019language} model is used to encode the constraints. As GPT-2 is more language-oriented, we input the natural language version of ASP constraints (while experimenting with ASP-form constraints to assess their impact on performance).

The problem is formulated as a multi-label classification problem where the output is one or more of the following $17$ labels - \rmfamily{\{\textit{red, blue, green, yellow, cyan, brown, gray, purple, rubber, metal, large, small, medium, cone, cube, cylinder, sphere}\}}.  Hence, the three encodings are passed to a multi-label classifier (feed-forward network) which is the only module of the whole model that is trained from scratch. The classifier is trained with a weighted binary cross entropy loss function \cite{ho2019real} that gives more penalty to the wrong prediction of minority class (as most of the labels in the output are 0, except for the ones in the answer - a false negative is given more penalty). For each of the $17$ labels, the weighted cross entropy loss is thus defined as below:
\begin{equation*}
\label{eq:1}
\small
WCE\left( {y,\hat y} \right) = - \left( {\beta y\log (\hat y) + (1 - y)\log (1 - \hat y)} \right)    \tag{1}
\end{equation*}
$\beta$ is the weight (is set $>1$ to penalize false negatives)\footnote{The results in Section~\ref{sec:results} are for $\beta=5$.}, $y$ is the ground truth, $\hat y$ is the prediction. 

\normalsize

\subsubsection{Neuro-Symbolic Visual Question Answering}\label{ns-vqa}
The architecture for the neuro-symbolic approach to solving CLEVR-POC task is shown in Figure \ref{fig:NSVQA}. The idea is to convert both the image and the question into a unified space as in CLIP, with the difference that this space is symbolic (scene graph and question in ASP). The architecture is based on the state-of-the-art neuro-symbolic approach on the CLEVR dataset, NS-VQA \cite{yi2018neural} and will be used here to study aspects of RQ2. We modify this architecture to include an ASP solver that takes as input - the scene in ASP, the question in ASP, and the environment constraints in ASP to derive the answer to the question. 

The question parser, (a Bidirectional Long Short Term Memory (BiLSTM) sequence to sequence model) is trained as in NS-VQA using REINFORCE - the reward is positive if the ASP program generated by the parser results in the correct answer, else it is $0$. The question parser is initially pre-trained in a fully supervised way with a small sample of (question, ASP program) pairs. 


The image perception network in NS-VQA is based on Detectron \cite{girshick2018detectron} and it was trained independently of the question parser in a supervised way. The ASP solver used is the same as the one used during the dataset generation phase.

\subsubsection{LLMs for solving CLEVR-POC}
LLMs are leveraged in two ways for solving a reasoning task like CLEVR-POC.

\noindent \textbf{LLM as question parser in NS-VQA}: In this approach, we use LLM as a question parser - converting the question into a semantic representation like ASP. The image is converted to a scene graph as done in NS-VQA. Both semantic representations are then passed on to a formal reasoner like an ASP solver to derive solutions consistent with the constraints.

\noindent \textbf{Stand-alone LLM}: The second approach is to provide both the image description and the question along with the constraints (in NL) as input to LLM and generate as a response the consistent solutions. We, here, assume as done in NS-VQA that the scene graphs are accurate, as our focus is on evaluating LLMs' ability to perform symbolic reasoning. CLEVR-POC, a synthetic dataset where environment-specific knowledge is not fixed, can assess LLMs' symbolic reasoning ability without data contamination (where the dataset becomes unusable once it has been exploited).

The LLM used in the experiments is GPT-4 \cite{openai2023gpt4} (See Appendix C for details about prompts used).

\begin{table*}
\begin{subtable}{\textwidth}
\scriptsize
\centering
\begin{tabular}{|l|c|c|c|c|c|c|}
\cline{1-7}
{\bf Dataset} &{\bf NS-VQA (BiLSTM)}& {\bf NS-VQA (GPT-4)}&{\bf CLIP-ASP} & {\bf CLIP-NL} & {\bf CLIP (no knowledge)} & {\bf GPT-4}    \\
\cline{1-7}
2000 &  0.0200 & \textbf{0.9250} & 0.0350 & 0.0600 & 0.0500  & 0.4626\\
6000 &  0.1516 & \textbf{0.9550} & 0.1500 & 0.1700  & 0.1183 & -\\
12000 &  0.2308 & \textbf{0.9441} & 0.1800 & 0.2283 &0.1483  &  -\\
\cline{1-7}
\end{tabular}
\caption{Exact answer accuracies of CLIP, NS-VQA and GPT-4 models on CLEVR-POC.}
\label{tab:res1}
\end{subtable}

\begin{subtable}{\textwidth}
\scriptsize
\centering
\begin{tabular}{|l|c|c|c|c|c|c|}
\cline{1-7}
{\bf Dataset} & {\bf NS-VQA (Bi-LSTM)} & {\bf NS-VQA (GPT-4)} & {\bf CLIP-ASP} &{\bf CLIP-NL} & {\bf CLIP (no knowledge)} & {\bf GPT-4}   \\
\cline{1-7}
2000 & 0.0591 & \textbf{0.9287} & 0.1000 & 0.1557 & 0.1412 & 0.5164\\
6000 & 0.3602 & \textbf{0.9578} & 0.3100 & 0.3403 & 0.2447 & - \\
12000 & 0.4331 & \textbf{0.9496} & 0.3600 & 0.4465 & 0.2912 & -\\
\cline{1-7}
\end{tabular}
\caption{Jaccard Index of CLIP, NS-VQA and GPT-4 models on CLEVR-POC}
\label{tab:res2}
\end{subtable}
\caption{Exact accuracies and Jaccard index scores of NS-VQA with BiLSTM and GPT-4 as question parsers, CLIP and GPT-4 on CLEVR-POC. CLIP-NL and CLIP-ASP take constraints in natural language and ASP, respectively. CLIP (no knowledge) is the performance of CLIP without constraints. }
\label{tab:res}
\end{table*}

\begin{table*}
\small
\centering
\begin{tabular}{|c|c|c|c|}
\hline
\multicolumn{1}{|l|}{\textbf{Sample Size}} & \multicolumn{1}{l|}{\textbf{PA (after pre-training)}} & \multicolumn{1}{l|}{\textbf{PA (after REINFORCE)}} & \multicolumn{1}{l|}{\textbf{PA (GPT-4)}} \\ \hline
\textbf{28 (prompt size)}                                & -                                                     & -                                                  & \textbf{0.9250}                            \\
$\approx$ 200                                        & 0.0512                                                & 0                                                  & -                                        \\
$\approx$ 1000                                       & 0.4487                                                & 0.0366                                             & -                                        \\
$\approx$ 2000                                       & 0.5043                                                & 0                                                  & -          \\                             

\cline{1-4}
\end{tabular}
\caption{Drop of program accuracies (PA) after REINFORCE and the performance of GPT-4 provided with just 28 (question, ASP program) pairs as prompt.}
\label{tab:res3}
\end{table*}

\subsection{Evaluation}
Let $A$ = \rmfamily{$\{$a1, a2,..$\}$} denote the set of values in the actual answer and $P$ = \rmfamily{$\{$p1, p2,..$\}$} denote the predicted answer set. We evaluate the performance of the two approaches on CLEVR-POC using the two metrics based on accuracy. 
    
\textbf{Exact Accuracy} checks whether the prediction made is exactly accurate, i.e.,  $A$ is exactly equal to $P$.    

    \begin{equation*}
    \small
    Exact\_Accuracy(A, P) =
    \begin{cases}
        1 & \text{if } x \in A \iff x \in P \\
        0 & \text{otherwise}
    \end{cases}
    \tag{2}
    \end{equation*}
    
    \textbf{Jaccard Index} computes the similarity between the actual answer and predicted answer sets as: 
    \begin{equation*} 
    \small
    Jaccard\_Index\left(A, P \right) = \frac{|A \cap P|}{|A \cup P|}\tag{3}\end{equation*}
    The value of Jaccard\_Index is between 0 (no common elements) and 1 (exact match). It gives some credit for partially correct answers as well.   
   

\subsection{Results}
\label{sec:results}
Tables \ref{tab:res1} and \ref{tab:res2} show the results for exact and partial answer accuracies respectively for NS-VQA, CLIP-based models, and stand-alone GPT-4 on CLEVR-POC. While NS-VQA (BiLSTM) uses a BiLSTM trained from scratch as the question parser, NS-VQA (GPT-4) uses pre-trained GPT-4 as the question parser. We experimented with varying dataset sizes - $2000$, $6000$, and then $12000$ instances. \footnote{The models are trained on Intel® CoreTM i7-12700K, 32GB RAM, and an NVIDIA GeForce RTX 3080 Ti for training.} It can be seen that with a multifold increase in the dataset size, there is an improvement in the answer accuracy, but the performance is not satisfactory.

\textbf{RQ1 - CLIP-based model analysis}: Since the question is not about some object in the scene, and the set of constraints to be satisfied is also not fixed across the instances in the dataset, it is challenging to learn a mapping from the three inputs (the incomplete scene, the natural language question, and the constraints) to the output set of plausible values. Table \ref{tab:res} shows three sets of results for CLIP. The columns CLIP-NL and CLIP-ASP correspond to instances of CLIP where the constraints are given in natural language and ASP respectively. It should be noted that CLIP-NL performs better than CLIP-ASP, suggesting that representing symbolic knowledge in natural language may be ideal while incorporating knowledge into neural frameworks for QA. The performance of CLIP on CLEVR-POC when no external knowledge is provided is shown in the column CLIP (no knowledge). Although without the external knowledge CLIP's performance drops, there is not much of a difference indicating that we need to consider better techniques for incorporating such symbolic constraints into neural frameworks. This points us toward existing neuro-symbolic frameworks. 

\textbf{RQ2 - NS-VQA analysis}: While neural models failed in symbolic reasoning and incorporating symbolic knowledge into the network, it can be seen that the major challenge faced by neuro-symbolic architectures lies not in reasoning but in mapping image or question to a symbolic representation in the absence of ground truth semantic representations.  In our experiments, we focus on language perception while assuming 100\% accuracy in image perception. Tackling both perceptions simultaneously is even more formidable without access to ground truth representations. Hence, the poor performance of NS-VQA (see column NS-VQA (BiLSTM) in Tables \ref{tab:res1} and \ref{tab:res2}) can be solely attributed to the failure of REINFORCE in learning accurate ASP programs.  As mentioned in Section \ref{ns-vqa}, a BiLSTM is initially pre-trained in a supervised fashion with a few examples. We experimented by varying the number of examples provided for pre-training. Table \ref{tab:res3} shows the program accuarcy after pre-training with around $200$, $1000$ and $2000$ pairs of $<$question, ASP program$>$. When these pre-trained models are further trained with REINFORCE, there is a drastic drop in the program accuracy as the focus of the REINFORCE algorithm is on coming up with the correct answer independent of the program's accuracy. This fall is observed even with the original CLEVR dataset. The chances of deriving the correct answer even with a wrong program by a fluke are higher in the case of CLEVR compared to CLEVR-POC considering the larger solution space of CLEVR-POC (see Section \ref{quest_repr}). REINFORCE clearly fails to learn ASP programs through weak supervision even when it initiates its training from a proficient model. 

\noindent{\textbf{RQ3 - LLM Analysis}}: In the first experiment we used GPT-4 as a question parser. The BiLSTM-based question parser of NS-VQA is replaced with GPT-4 (the results are shown in column NS-VQA(GPT-4) in Tables \ref{tab:res1} and \ref{tab:res2}). The model is provided with just 28 (question, ASP program) pairs of examples as prompts. GPT-4 with no fine tuning was able to accurately predict the equivalent ASP programs.  

The stand-alone GPT-4 approach gave less than 50\% exact accuracy. The evidence indicates that employing GPT-4 as a question parser to translate the question into an ASP program and subsequently utilizing an ASP reasoning engine leads to better results compared to placing the entire burden of symbolic reasoning on GPT-4. It should also be noted that GPT-4 with no data-specific training performed better than CLIP and NS-VQA (BiLSTM). There is still room for improvement with some fine-tuning.

\section{Discussion}

We now discuss important challenges that our dataset and work point to.

\noindent \textbf{Reasoning and LLM}: The experiments showed that the direct application of LLMs is not a good solution for such reasoning-intensive tasks. \cite{mahowald2023dissociating} also discusses the limitations of LLMs in formal reasoning tasks. Our experiments showed that a more appropriate approach to harnessing LLMs involved relieving them of the task of symbolic reasoning and instead employing them for generating symbolic representations. Progressing even further entails discovering mechanisms for seamlessly incorporating specific knowledge into LLMs and generating responses that are consistent with this knowledge.

\noindent \textbf{Symbolic knowledge in visual perception network}: Although the focus of this paper was on language and reasoning, it may be noted that knowledge in the form of constraints in CLEVR-POC can play a significant role during image perception as it can provide hints on what can or cannot be in the image. This is a form of weak supervision which is also required in the absence of ground truth scene graphs to accelerate the learning process. Developing neuro-symbolic models with a stronger feedback mechanism for visual perception, such as DeepProbLog \cite{manhaeve2018deepproblog}, NeurASP \cite{yang2020neurasp}, Semantic-Loss~\cite{xu2018semantic} and LTN~\cite{serafini2016logic}), would help in faster convergence. The aforementioned frameworks, however, cannot still be applied to VQA tasks due to scalability issues. 


\section{Conclusion}
Humans often have to interact with the partially observable environment. In light of the need to deal with the inherent uncertainty in knowledge-rich real-world scenarios, this work aimed to establish a benchmark for evaluating reasoning-intensive VQA in partially observable environments. Applying the benchmark to stand-alone LLMs and other vision-language models yielded disappointing results due to their inability to perform symbolic reasoning. We also demonstrated that combining LLM with a visual perception network and a formal reasoner produced positive results. 

Future directions involve developing visual perception networks with knowledge-guided supervision, enhancing LLMs' reasoning capabilities, and moving CLEVR-POC to an embodied setup like vision language navigation.

\section{Acknowledgements}
This research was conducted during the authors' tenure at Örebro University, Sweden and was financially supported by the Wallenberg AI, Autonomous Systems, and Software Program (WASP).

\nocite{*}
\section{Bibliographical References}\label{sec:reference}

\bibliographystyle{lrec_natbib}
\bibliography{references}



\clearpage

\appendix
\section{An example from CLEVR-POC}
\label{appendixA}
\begin{figure}
\centering
\includegraphics[width=0.45\textwidth, height=4cm]{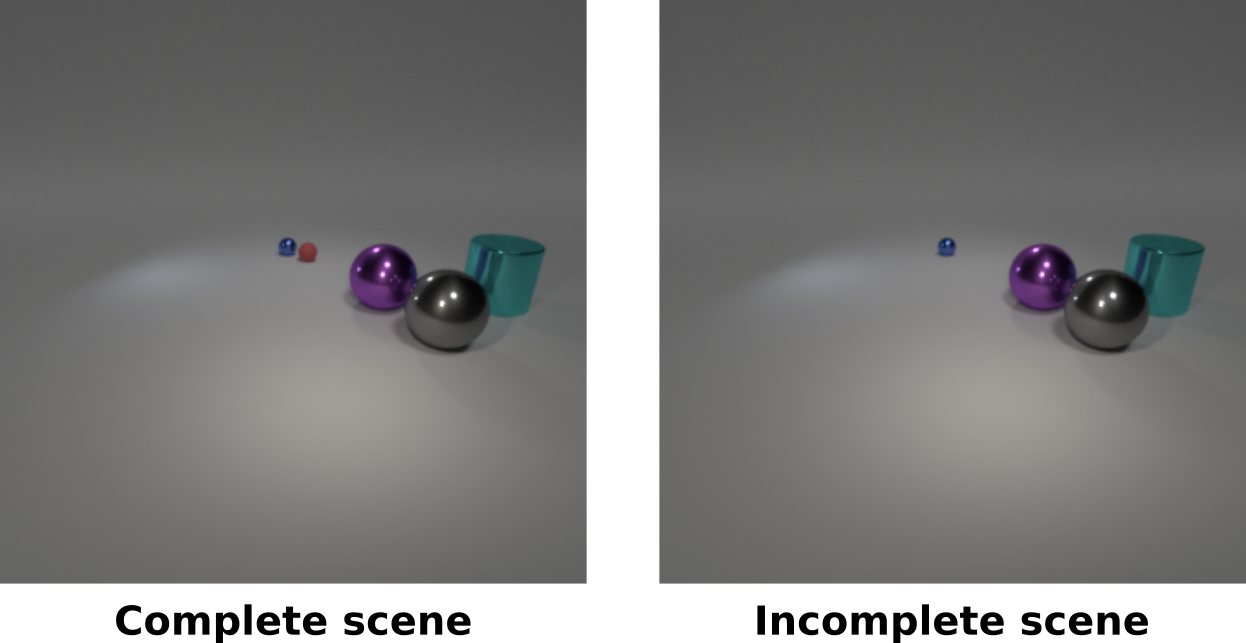}
\caption{A complete and incomplete scene from CLEVR-POC}
\label{fig:CLIP-appendix}
\end{figure}

\noindent \textbf{Complete and incomplete scene}:
Figure \ref{fig:CLIP-appendix} is an example of a complete scene and the incomplete scene generated from it by hiding the \textit{small red rubber sphere}. 

\noindent \textbf{Environment}
Every scene is generated such that it satisfies the constraints in an environment. The following are the general rules shared by all environments in CLEVR-POC. 

\begin{mdframed}

\fontsize{7pt}{8pt}\selectfont
\begin{verbatim}
 1. property(color, gray). property(color, red).
 2. property(color, blue). property(color, green).  
 3. property(color, brown). property(color, purple).  
 4. property(color, cyan). property(color, yellow).  
 5. property(shape, cube). property(shape, cylinder).  
 6. property(shape, sphere). property(shape, cone).  
 7. property(size, small). property(size, medium).  
 8. property(size, large).  
 9. property(material, rubber). 
    property(material, metal).  
10. region(0). region(1). region(2). region(3).  
11. right_R(0, 0). right_R(0, 1). right_R(0, 2). 
    right_R(0, 3).  
12. right_R(1, 1). right_R(1, 3).  
13. right_R(2, 0). right_R(2, 1). right_R(2, 2). 
    right_R(2, 3).  
14. right_R(3, 1). right_R(3, 3).  
15. left_R(R1, R2) :- right_R(R2, R1).  
16. front_R(0, 0). front_R(0, 1). front_R(0, 2). 
    front_R(0, 3).  
17. front_R(1, 0). front_R(1, 1). front_R(1, 2). 
    front_R(1, 3).  
18. front_R(2, 2). front_R(2, 3).  
19. front_R(3, 2). front_R(3, 3).  
20. behind_R(R1, R2) :- front_R(R2, R1).  
21. sameProperty(X1, X2, P) :- hasProperty(X1,P,V),}  
22. hasProperty(X2,P,V), X1!=X2.  
23. same_color(X,Y):- sameProperty(X, Y, color).                
24. same_size(X,Y):- sameProperty(X, Y, size).  
25. same_shape(X,Y):- sameProperty(X, Y, shape).  
26. same_material(X,Y):- sameProperty(X, Y, material).  
27. 1{hasProperty(X, color, V) : 
28.            property(color, V)}1 :-  object(X).  
29. 1{hasProperty(X, material, V) : 
30.            property(material, V)}1 :- object(X).  
31. 1{hasProperty(X, shape, V) : 
32.            property(shape, V)}1 :- object(X).  
33. 1{hasProperty(X, size, V) : 
34.            property(size, V)}1 :- object(X).                   
35.1{at(X, R): region(R)}1 :- object(X).  
36.:- sameProperty(X1, X2, color),  
37.   sameProperty(X1, X2, material),   
38.   sameProperty(X1, X2, size)}, 
39.   sameProperty(X1, X2, shape),  
40.   object(X1), object(X2), X1!=X2.  
41.exceed_region_capacity(R) :-
42.   #count{X: object(X), at(X, R)} >= 4, region(R).              
43:- exceed_region_capacity(_).
                  
\end{verbatim}
\end{mdframed}
Environment's general rules in natural language:

\begin{mdframed}

\fontsize{7pt}{8pt}\selectfont
\begin{verbatim}
1-9. Objects must have 4 properties. They are color,
shape, size, and material.
                  
1-4. Objects can be in one of the 8 colors. It can 
be gray, or red, or blue, or green, or brown, 
or purple, or cyan,  or yellow.

5-6. Objects can be in one of the 4 shapes. 
It can be cube, or a cylinder, or a sphere or cone.

7-8. Objects can be in one of the 3 sizes. 
It can be small, medium, or large.

9. Objects can be in one of the 2 materials. 
It can be rubber or metal.

10. The scene is divided into 4 regions. 
They are named 0, 1, 2, 3.
                  
11. If there are two objects, the first object is 
located  in region 0 and the second object is to 
the right of the  first object, then the location 
of the second object is either in region 0, 1, or 
2, or 3.

12. If there are two objects, the first object is
located in region 1 and the second object is to
the right of the first object, then the location
of the second object is either in region 1, or 3.

13. If there are two objects, the first object is
located in region 2 and the second object is to 
the right of the first object, then the location
of the second object is either in region 0, 1, 2, or 3.

14. If there are two objects, the first object is
located in region 3 and the second object is to 
the right of the first object, then the location
of the second object is either in region 1, or 3.

15. If there are two objects, the first object is
to the right of the second object, then the second
the object is to the left of the first object.

16. If there are two objects, the first object is
located in region 0 and the second object is in 
front of the first object, then the location of 
the second  object is either  in region 0, 1, or 
2, or 3.

17. If there are two objects, the first object is
located in region 1 and the second object is in 
front of the first object, then the location of 
the second object is either in region 0, 1, or 
2, or 3.

18. If there are two objects, the first object is 
located in region 2 and the second object is in 
front of the first object, then the location of 
second object is either in region 2, or 3.

19. If there are two objects, the first object is 
located in region 3 and the second object is in 
front of the first object, then the location of 
the second object is either in region 2, or 3.

20. If there are two objects, the first object is 
in front of the second object, then the second 
the object is behind the first object.
                                    
27-28. Every object must be assigned exactly one 
value for color.

29-30. Every object must be assigned exactly one 
value for the material.

31-32. Every object must be assigned exactly one 
value for shape.

33-34. Every object must be assigned exactly one 
value for size.

35. Every object must be assigned exactly one value 
for region.

36-40. Two different objects cannot have the same values 
for all the 4 properties.
                  
41-43. Every region can have at most 3 objects.
\end{verbatim}
\end{mdframed}
The following constraints in ASP represent the specific environment to which the scene in Figure \ref{fig:CLIP-appendix} belongs.

\begin{mdframed}  

\fontsize{7pt}{8pt}\selectfont
\begin{verbatim}
44. object(0..4).
45. :- object(X), at(X, 0), 
                hasProperty(X, size, large).
46. :- object(X), at(X, 0), 
                hasProperty(X, shape, cylinder).
47. :- object(X), at(X, 0), 
                hasProperty(X, shape, cone).
48. :- object(X), at(X, 1), 
                hasProperty(X, size, small).
49. :- object(X), at(X, 1), 
                hasProperty(X, shape, cone).
50. :- object(X), at(X, 1), 
                hasProperty(X, material, rubber).
51. :- object(X), at(X, 1), 
                hasProperty(X, shape, cube).
52. :- object(X), at(X, 2), 
                not hasProperty(X, size, medium).
53. :- object(X), at(X, 2), 
                not hasProperty(X, material, metal).
54. :- object(X), at(X, 2), 
                hasProperty(X, material, rubber).
55. :- object(X), at(X, 2), 
                hasProperty(X, shape, sphere).
56. :- object(X), at(X, 2), 
                hasProperty(X, shape, cube).
57. :- object(X), at(X, 3), 
                hasProperty(X, size, small).
58  :- object(X), at(X, 3), 
                not hasProperty(X, material, metal),
59. not hasProperty(X, color, blue).
60. :- #count{X1, X2: sameProperty(X1, X2, shape),
61.     object(X1), object(X2), at(X1, 3), at(X2, 2),
62.     hasProperty(X1, color, yellow),
63.     hasProperty(X2, color, yellow)} >= 4.
64. :- #count{X1, X2: sameProperty(X1, X2, color), 
65.     object(X1), object(X2),
66.     at(X1, 0), at(X2, 3)} >= 2.
\end{verbatim}
\end{mdframed}

The following is a natural language interpretation of each line of the preceding rules.
\begin{mdframed}
\fontsize{7pt}{8pt}\selectfont
\begin{verbatim}
44. There are 5 objects in the scene.
45. There are no large size objects in region 0. 
46. There are no cylinder shape objects in region 0. 
47. There are no cone shape objects in region 0. 
48. There are no small size objects in region 1. 
49. There are no cone shape objects in region 1. 
50. There are no rubber material objects in region 1. 
51. There are no cube shape objects in region 1. 
52. All objects in region 2 have medium size. 
53. All objects in region 2 have metal material. 
54. There are no rubber material objects in region 2. 
55. There are no sphere shape objects in region 2. 
56. There are no cube shape objects in region 2. 
57. There are no small size objects in region 3. 
58-59. All objects in region 3 have either metal 
material or blue color. 
60-63. There are at most 1 pairs of color yellow 
objects with the same shape in regions 3 and 2 
together. 
64-66. There are at most 0 pairs of objects with the 
same color in regions 0 and 3 together.

\end{verbatim}    
\end{mdframed}

\noindent \textbf{Question:} 
For each given incomplete scene, we generate one question about (any property) of the missing object. The following is the \textbf{question in natural language} that is associated with the incomplete scene in Figure \ref{fig:CLIP-appendix}:
    
        \vspace{0.2cm}
         \fbox{%
          \parbox{0.45\textwidth}{
           \textit{There is another red rubber object that is the same shape as the big purple object; what size is it?}
          }%
        }
        \vspace{0.3cm}
    
The following is the same \textbf{question represented in ASP}:

\begin{mdframed}
   
\fontsize{7pt}{8pt}\selectfont
\begin{verbatim}
 1.  missing(Q)  :-  
 2.         hasProperty(X,size,Q),
 3.         hasProperty(X,material,rubber),
 4.         hasProperty(X,color,red),
 5.         hasProperty(Y,color,purple),
 6.         hasProperty(Y,size,large),
 7.         X!=Y,
 8.         same_shape(Y,X).            
\end{verbatim} 
\end{mdframed}

\noindent \textbf{Answer set:} The answer set for the above question that satisfies the constraints in the specified environment is:

\begingroup
\fontsize{8pt}{9pt}\selectfont
\begin{verbatim}    
{small, medium}
\end{verbatim} 
\endgroup       

\noindent \textbf{Reasoning Steps:} The reasoning involved in deriving the answer set from the question, the incomplete scene, and the constraints in the specified environment is given below.
        \begin{itemize}
            \item \vspace{0.2cm} 
            Interpreting each line of the question:

\begingroup
\fontsize{8pt}{9pt}\selectfont
\begin{verbatim}
1. What are the possible values for Q such that:
2. Q is size of the missing object,
3. the missing object's material is rubber,
4. the missing object's color is red,
5. the reference object's color is purple,
6. the reference object's size is large,
7. the missing object is not equal to the 
   reference object,
8. the missing object's shape = the 
   reference object's shape.      
\end{verbatim} 
\endgroup 
            \item Inferring the missing object's properties:\\
            from the scene graph: =$>$

\begin{figure*}[t]
    \centering
    \includegraphics[width=\textwidth]{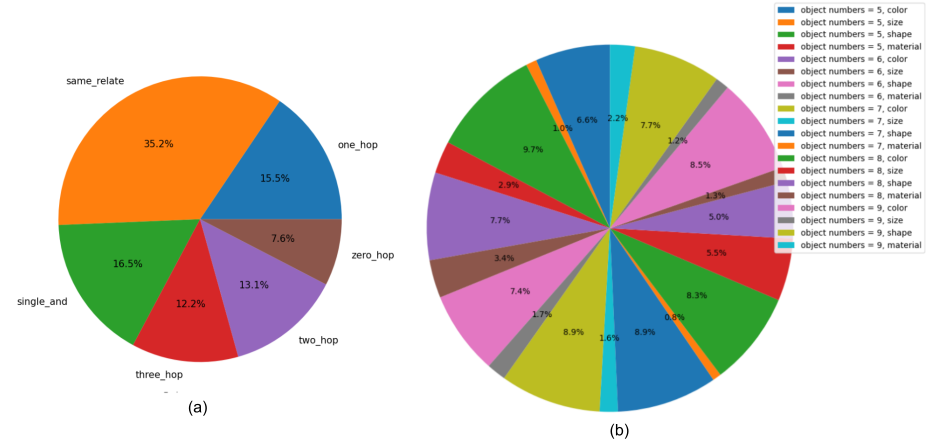}
    \caption{(a) Question templates distribution (b) Distribution of query attributes with object counts between 5 to 9.}
    \label{fig:my_label1}
\end{figure*}

\begingroup
\fontsize{8pt}{9pt}\selectfont
\begin{verbatim}
8. the reference object's shape is a sphere.        
9. => The missing object's shape is also a sphere.
10. => The missing object is a red rubber sphere.
\end{verbatim}
\endgroup 
            \item Inferring the missing object's possible \textbf{regions} based on the rules listed as the \textbf{Environment's constraints}:\\\\
            Among the four regions:
\begingroup
\fontsize{8pt}{9pt}\selectfont
\begin{verbatim}
A red rubber sphere CAN be located at region 0, 
as none of the constraints in lines 45-47 is 
violated.

A red rubber sphere CAN'T be located in region 
1, as it violates the constraint about the 
material at line 50.

A red rubber sphere CAN'T be located in region 
2, as it violates the constraints in lines 53,
54, and 55.

A red rubber sphere CAN'T be located in region
3, as it violates the constraints in lines 58-59.

=> The missing red rubber sphere is located at 
region 0.
\end{verbatim}
\endgroup

                \item Inferring the possible answer set for the property of interest w.r.t the inferred location of the missing object:

\begingroup
\fontsize{8pt}{9pt}\selectfont
\begin{verbatim}
There are 3 possible values for the size 
property:
small, medium, large.

The environment constraint at line 45 discards 
the large size for region 0.

=> The possible answer set for Q is: 
small, medium.   
\end{verbatim}
\endgroup
        \end{itemize}

\section{Dataset Statistics}

\begin{figure*}[t]
    \centering
    \includegraphics[width=\textwidth]{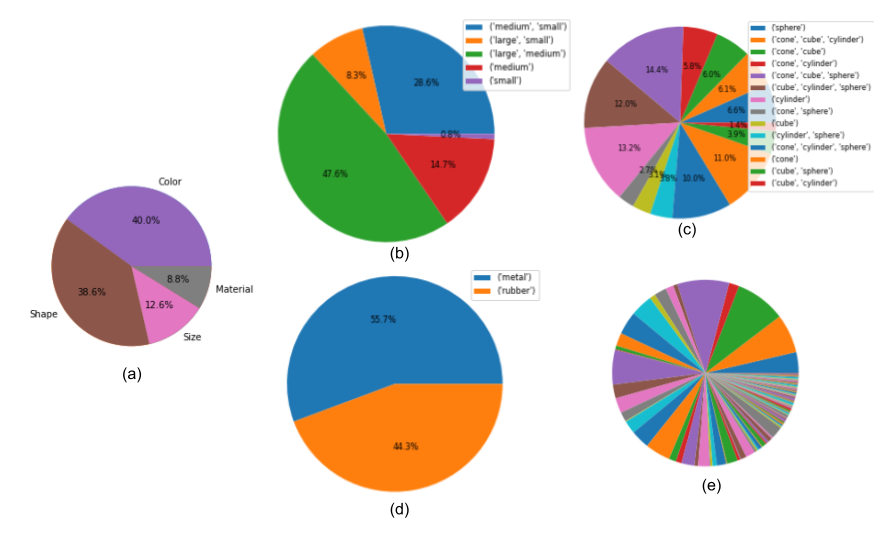}
    \caption{(a) Distribution of question types. (b) Distribution of solutions for questions with query attribute size. (c) Distribution of solutions for questions with query attribute material. (d) Distribution of solutions for questions with query attribute shape. (e) Distribution of solutions for questions with query attribute color. Since the solution space of these questions is larger ($>100$), it is not listed here.}
    \label{fig:my_label2}
\end{figure*}

\noindent \textbf{Distribution across question templates}: Figure \ref{fig:my_label1} (a) shows the distribution of questions across different question templates. Six templates present in the original CLEVR dataset are used in CLEVR-POC.

\noindent \textbf{Distribution of query attributes with number of objects in the scene}: Figure \ref{fig:my_label1} (b) shows the distribution of questions of a specific type based on the number of objects in the scene. 

\noindent \textbf{Distribution across question types}: The type of question asked depends on the attribute of the object that is being inquired about. The generation process enables the user to have control over this distribution. For instance, when generating the specific dataset that was used in the experiments, we established the following criteria: $40$\% of the questions pertain to the color attribute, another $40$\% focus on the shape attribute, $10$\% address the size attribute, and the remaining $10$\% relate to the material attribute. We made this selection based on the observation that attributes like color and shape encompasses a larger set of values (8 values for color and 4 for shape) in comparison to material (which has just two values). Consequently, the solution space for questions centered around color is more extensive than that for material, resulting in a more diverse solution space for the dataset.  Figure \ref{fig:my_label2} (a) displays the question type distribution of the dataset generated based on this setting.

\noindent \textbf{Distribution across solutions}:
Figure \ref{fig:my_label2} (b), (c), (d), and (e) illustrate the distribution of potential solutions for various question types: size, shape, material, and color, respectively. We aim for a balanced distribution, avoiding a situation where the majority of questions lead to the same set of answers. For instance, when a question pertains to the size of an object, its possible solutions could be one of \{large, medium\} or \{large, small\}, or \{small, medium\} or \{large\}, or \{medium\} or \{small\} as depicted in Figure \ref{fig:my_label2} (b).  Since the possible solutions for questions with query attribute \textit{color} are large (as \textit{color} can take 8 values), the entire space is not listed in Figure \ref{fig:my_label2}(e). However, it can be seen that the distribution is not favoring any specific solution.

\section{Prompts for Language Model}

\subsection{Stand-alone GPT-4 to solve CLEVR-POC}

The format of the prompt provided to GPT-4 when employing it to solve CLEVR-POC is shown below. The prompt contains the task description, the scene description, the constraints or knowledge associated with the scene, the question about the scene, and the answer. The prompt contains two such examples.  

\begin{mdframed}
\begingroup
\fontsize{8.5pt}{7.5pt}\selectfont

\textcolor{blue} {\textbf{Task description}: You are a helpful assistant who answers questions about hidden objects based on scene description and the constraints in the scene. The scene graph is in JSON format with the following keys. The key objects contain a list of objects present in the scene. Each object has various attributes like material, color, shape, size, and region. The key relationships hold information about the spatial relationships between objects in the scene. It contains sub-fields like "front," "right," "left," "behind," etc., each associated with a list of object indices representing objects that have that specific relationship with another object. For example, relationships["front"][0] refers to the objects that are in front of the object at index 0. }

\noindent \textcolor{mypurple} {\textbf{Scene Observed}:
The following is the scene graph:\\}
\endgroup

\begin{lstlisting}[language=Python, style=plaintextstyle]
{'objects': [
 {'material': 'metal',  'color': 'red', 
  'size': 'medium',  'region': '0', 
  'shape': 'cube'
 },
 {'material': 'metal', 'color': 'gray', 
  'size': 'medium', 'region': '3', 
  'shape': 'sphere'
 }, 
 {'material': 'metal', 'color': 'brown', 
  'size': 'medium', 'region': '1', 
  'shape': 'sphere'
 }, 
 {'material': 'rubber', 'color': 'gray',
  'size': 'medium', 'region': '3',
  'shape': 'sphere'
 }, 
 {'material': 'metal', 'color': 'red', 
  'size': 'medium', 'region': '0', 
  'shape': 'sphere'
 }, 
 {'material': 'rubber', 'color': 'red', 
  'size': 'medium', 'region': '2', 
  'shape': 'sphere'
 }
],
 'relationships': 
  {'left': [[4], [0, 2, 4, 5], [0, 4, 5],  [0, 1, 2, 4, 5], [], [0, 4]], 
   'front': [[1, 3, 4, 5], [5], [0, 1, 3, 4, 5], [1, 5], [1, 3, 5], []], 
   'behind': [[2], [0, 2, 3, 4], [], [0, 2, 4], [0, 2], [0, 1, 2, 3, 4]], 
   'right': [[1, 2, 3, 5], [3], [1, 3], [], [0, 1, 2, 3, 5], [1, 2, 3]]
  } 
}
\end{lstlisting}

\begingroup
\fontsize{8.5pt}{7.5pt}\selectfont
\noindent \textcolor{mypurple} {\textbf{Constraints}: The scene contains several visible objects, and has one additional object that is hidden. Objects must have 4 properties. They are color, shape, size, and material. The scene must conform to the following constraints.} 
\endgroup
\begin{lstlisting}[style=plaintextstyle]

Objects can be in one of the 8 colors. It can be gray, or red, or blue, or green, or brown, or purple, or cyan, or yellow.

Objects can be in one of the 4 shapes. It can be a cube, cylinder, sphere, or cone.

Objects can be in one of the 3 sizes. It can be small, medium, or large.

Objects can be in one of the 2 materials. It can be rubber or metal.

The scene is divided into 4 regions. They are named 0, 1, 2, 3.

If there are two objects and the first object is located in region 0 and the second object is to the right of the first object, then the location of the second object is either in region 0, 1, 2, or 3.

If there are two objects and the first object is located in region 1 and the second object is to the right of the first object, then the location of the second object is either in region 1, or 3.

If there are two objects and the first object is located in region 2 and the second object is to the right of the first object, then the location of the second object is either in region 0, 1, 2, or 3.

If there are two objects, the first object is located in region 3 and the second object is to the right of the first object, then the location of the second object is either in region
1, or 3.

If there are two objects, the first object is to the right of the second object, then the second object is to the left of the first object.

If there are two objects, the first object is located in region 0 and the second object is in front of the first object, then the location of the second object is either in region 0, 1, 2, or 3.

If there are two objects, the first object is located in region 1 and the second object is in front of the first object, then the location of the second object is either in region 0, or 1, or 2, or 3.

If there are two objects, the first object is located in region 2 and the second object is in front of the first object, then the location of the second object is either in region 2, or 3.

If there are two objects, the first object is located in region 3 and the second object is in front of the first object, then the location of the second object is either in region 2, or 3.

If there are two objects, the first object is in front of the second object, then the second object is behind the first object.

Every object must be assigned exactly one value for color.

Every object must be assigned exactly one value for material.

Every object must be assigned exactly one value for shape.

Every object must be assigned exactly one value for size.

Every object must be assigned exactly one value for region.

Two different objects cannot have the same values for all the 4 properties.

Every region can have at most 3 objects.

There are 6 objects in the scene.

There are at least 1 pair of color red objects with the same size in regions 0 and 2 together.

There are no small-size objects in region 0.

There are no cone-shaped objects in region 0.

There are no purple color objects in region 0.

There are no blue color objects in region 0.

There are no cylinder shape objects in region 1.

There are no cyan color objects in region 1.

There are no rubber material objects in region 1.

There are at least 1 pair of material metal objects with the same size in regions 0 and 3 together.

There are no metal material objects in region 2.

There are no large-size objects in region 2.

There are at least 1 pair of size medium objects with the same shape in regions 1 and 3 together.

There are no red color objects in region 3.

There are no cube-shaped objects in region 3.

There are at least 1 pair of objects with the same material in regions 0 and 1 together.

There are at least 1 pair of color gray objects with the same size in regions 3 and 2 together.


                  
\end{lstlisting}
\begingroup
\fontsize{8.5pt}{7.5pt}\selectfont
\noindent \textcolor{mypurple} {\textbf{Question}: Answer the following question about the hidden object. The solution should satisfy the constraints. The other cylinder that is the same material as the medium red thing is what color?}

\noindent \textcolor{mypurple} {\textbf{Answer}:} \textcolor{teal} { \#\#\#\{Gray\}\#\#\#}
\endgroup

\end{mdframed}

\subsection{GPT-4 as Question Parser}
When we use GPT-4 to parse a question to its ASP equivalent, we give as prompt $28$ examples of questions in natural language to ASP representation. The prompt with just one Question-ASP pair is shown below. 

\begin{mdframed}
\begingroup
\fontsize{8.5pt}{7.5pt}\selectfont
\noindent \color{blue} {\textbf{Task description}: You are a helpful assistant that converts questions in English into ASP logic language.}
\\

\noindent \color{mypurple} {\textbf{Question}: What is the color of the cylinder to the right of the blue sphere? } 

\noindent \color{mypurple}{\textbf{ASP:}}
\endgroup

\begin{lstlisting}[language=Python, style=plaintextstyle]
### 
unknown(Q):-hasProperty(X, color, Q), 
             hasProperty(X, shape, cylinder), 
             hasProperty(X1, color, blue), 
             hasProperty(X1, shape, sphere), 
             right(X1, X).
###
\end{lstlisting}

\end{mdframed}

\end{document}